\documentclass[runningheads]{llncs}

\usepackage{eccv}

\usepackage{xcolor}
\usepackage{colortbl}
\usepackage{booktabs}
\usepackage{booktabs}   
\usepackage{multirow}   
\usepackage{array}      
\usepackage{amssymb}    
\usepackage{mathtools} 
\usepackage{xcolor}

\definecolor{PositiveGreen}{RGB}{0, 145, 0}  
\definecolor{NegativeRed}{RGB}{210, 0, 0}    

\newcommand{\up}[1]{\ensuremath{\mathrlap{_{\textcolor{PositiveGreen}{+#1}}}}}
\newcommand{\down}[1]{\ensuremath{\mathrlap{_{\textcolor{NegativeRed}{-#1}}}}}

\usepackage{eccvabbrv}

\usepackage{graphicx}
\usepackage{booktabs}

\usepackage[accsupp]{axessibility}  

\usepackage{hyperref}

\usepackage{orcidlink}

\begin{document}

\title{DreamerAD: Efficient Reinforcement Learning via Latent World Model for Autonomous Driving} 

\titlerunning{DreamerAD}

\author{
    Pengxuan Yang\inst{1,2,3}\thanks{Work done during internship at Chongqing Chang’an Technology Co., Ltd.} \and
    Yupeng Zheng\inst{1}\thanks{Project Leader} \and
    Zebin Xing\inst{1} \and
    Qichao Zhang\inst{1,4}$^{\dagger}$  \and \\
    Deheng Qian\inst{2}$^{\dagger}$ \and
    Linbo Wang\inst{1,} \and
    Yichen Zhang\inst{1}\and
    Shaoyu Guo\inst{1} \and \\
    Zhongpu Xia\inst{1} \and 
    Qiang Chen\inst{2} \and 
    Junyu Han\inst{2} \and
    Lingyun Xu\inst{2} \and
    Yifeng Pan\inst{2}\protect\footnotemark[2] \and 
    Dongbin Zhao\inst{1}
}

\authorrunning{P.~Yang et al.}

\institute{
    Institute of Automation, CAS \and
    Chongqing Chang’an Technology Co., Ltd \and
    School of Advanced Interdisciplinary Sciences, UCAS \and
    School of Artificial Intelligence, UCAS
}

\maketitle
{\let\thefootnote\relax\footnotetext{\hspace*{-0.8em}$^{\dagger}$ Corresponding author.}}
\begin{abstract}
We introduce DreamerAD, the first latent world model framework that enables efficient reinforcement learning for autonomous driving by compressing diffusion sampling from 100 steps to 1—achieving 80× speedup while maintaining visual interpretability.
Training RL policies on real-world driving data incurs prohibitive costs and safety risks. While existing pixel-level diffusion world models enable safe imagination-based training, they suffer from multi-step diffusion inference latency (2s/frame) that prevents high-frequency RL interaction.
Our approach leverages denoised latent features from video generation models through three key mechanisms: (1) shortcut forcing that reduces sampling complexity via recursive multi-resolution step compression, (2) an autoregressive dense reward model operating directly on latent representations for fine-grained credit assignment, and (3) Gaussian vocabulary sampling for GRPO that constrains exploration to physically plausible trajectories.
DreamerAD achieves 87.7 EPDMS on NavSim v2, establishing state-of-the-art performance and demonstrating that latent-space RL is effective for autonomous driving.

\end{abstract}

\begin{figure}[htbp]
    \centering
    
    \begin{subfigure}{\textwidth}
        \centering
        \includegraphics[width=\linewidth]{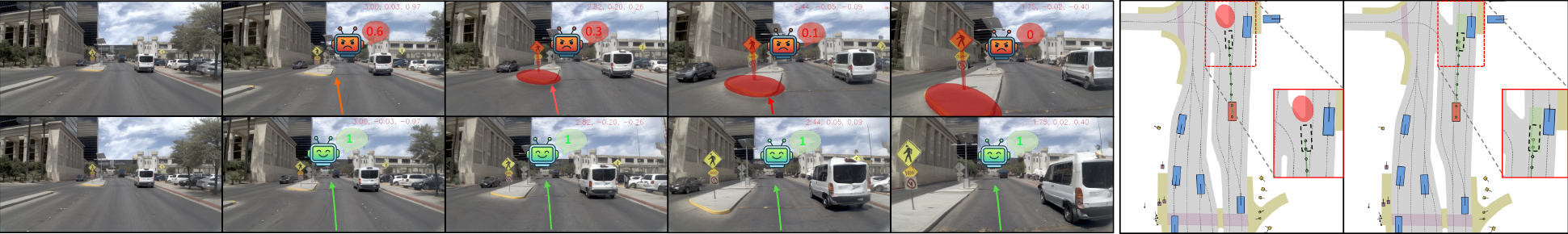}
        \caption{Scenario: Potential collision with the curb.}
        \label{fig:teaser_sub1}
    \end{subfigure}

    \begin{subfigure}{\textwidth}
        \centering
        \includegraphics[width=\linewidth]{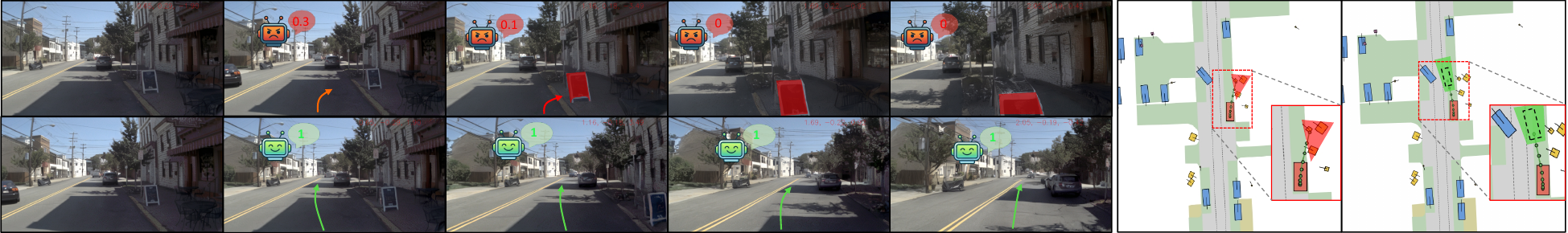}
        \caption{Scenario: collision with a roadside billboard.}
        \label{fig:teaser_sub2}
    \end{subfigure}

    \begin{subfigure}{\textwidth}
        \centering
        \includegraphics[width=\linewidth]{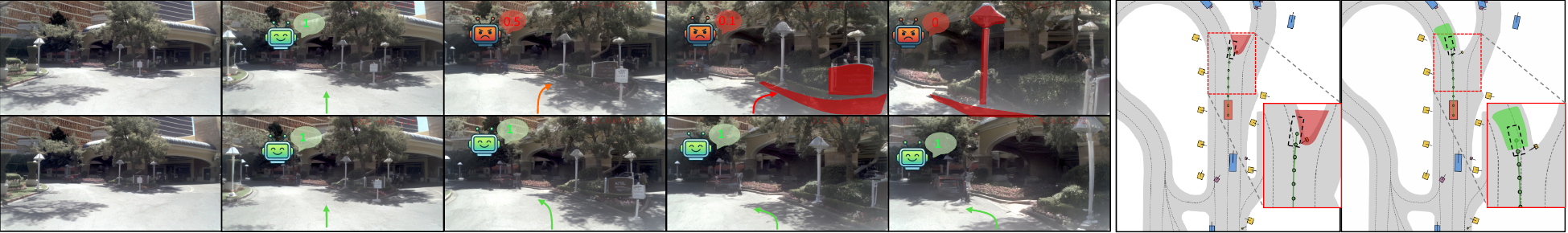}
        \caption{Scenario: collision with a street lamp.}
        \label{fig:teaser_sub3}
    \end{subfigure}

    \caption{\textbf{World model imagination training guided by diverse trajectories.} Each row shows a driving scenario where the world model imagines future outcomes for candidate trajectories. RGB sequences display predicted frames with reward model scores (red: collision risk, green: safe). BEV maps (right) visualize trajectories: hazardous paths (left, red-highlighted) versus safe alternatives (right, green-highlighted).} 
    \label{fig:teaser}
\end{figure}

\begin{figure}
    \centering
    \includegraphics[width=0.75\linewidth]{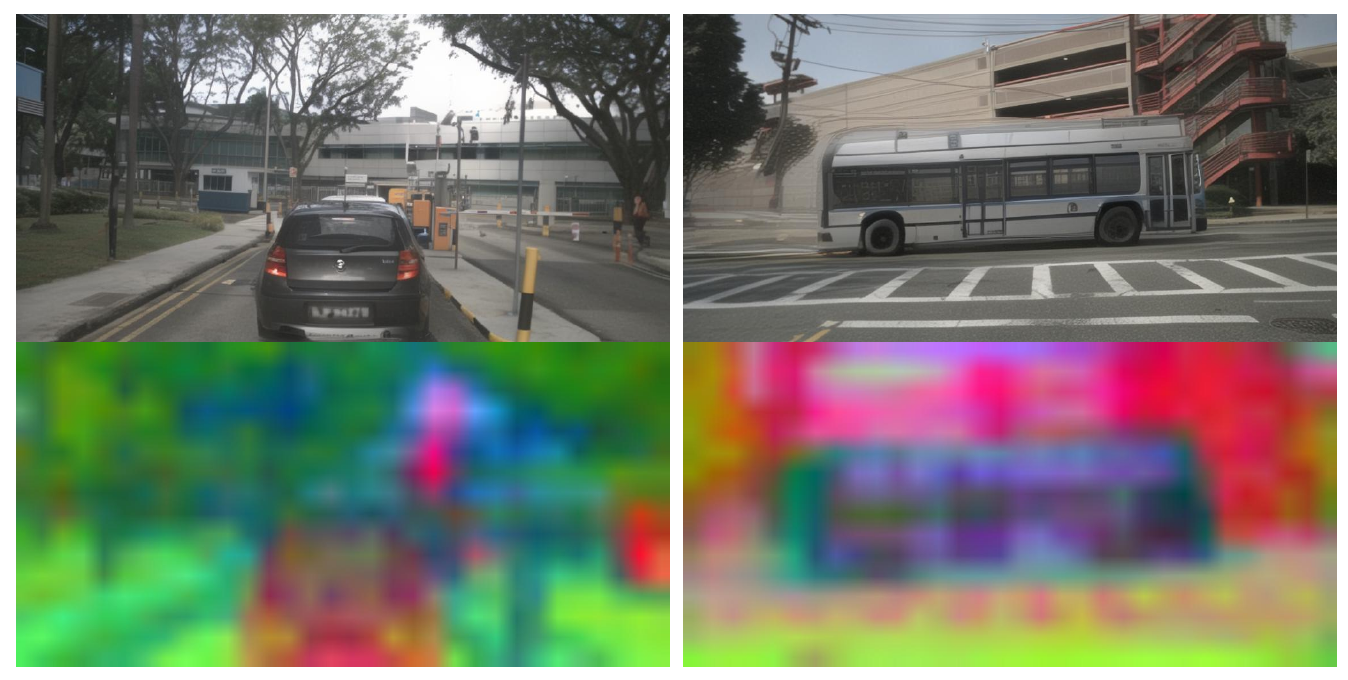}
    \caption{PCA visualization of denoised latent features, demonstrating strong spatial and semantic coherence.}
    \label{fig:pcb}
\end{figure}


\section{Introduction}

Reinforcement learning (RL) is widely recognized as an effective approach for addressing long-tail problems in autonomous driving, such as distribution shift and causal confusion. Training RL policies on real-world data incurs prohibitive trial-and-error costs and unacceptable safety risks. Conventional simulator-based methods introduce additional sim-to-real discrepancies. 
World models based on video generation offer a promising alternative by enabling imagination-based policy learning~\cite{jiang2026wovr, jiang2025world4rl}. However, existing pixel-level diffusion models\cite{zhang2025epona} face two critical bottlenecks: (1) multi-step sampling (100 steps) creates severe inference latency incompatible with RL's interaction demands, and (2) pixel-level objectives prioritize visual fidelity over the spatial and dynamic understanding crucial for driving safety.

To address these challenges, we propose DreamerAD, a latent world model framework that performs RL entirely within the latent imagination space of a video generation model. Our key insight is to leverage the denoised features from video generation models to construct a latent world model. Specifically, as shown in Fig.~\ref{fig:pcb}, we discover that the denoised latent features from Video DiT exhibit well-structured spatial information and semantic coherence. Building upon this latent world model, DreamerAD trains a reward model for RL training.
Concretely, we develop an inference-efficient world model by fine-tuning an autoregressive diffusion world model on the NavSim~\cite{im2024navsim} dataset with shortcut forcing, compressing world model sampling from 100 steps to a single step. This enables low-latency RL training in latent space while keeping the features losslessly decodable into high-fidelity RGB frames for rigorous interpretability.
Second, we construct an RL algorithm based on latent world model simulation to further enhance planning performance. Specifically, as shown in Fig.~\ref{fig:teaser}, we design a reward model that takes latent features conditioned on each action as "imagined inputs" to score each action step, providing dense quality assessment and credit assignment. Building upon the reward model, we propose a Gaussian vocabulary sampling-based GRPO optimization method that selects candidate trajectories from the neighborhood of the trajectory vocabulary based on Gaussian distributions. Compared to the random Gaussian point sampling in prior work~\cite{yang2025worldrft}, neighborhood vocabulary-based sampling ensures effective policy exploration, achieving more physically plausible and smoother trajectory planning.

We validate DreamerAD on the NavSim v2 closed-loop benchmark. DreamerAD achieves 87.7 EPDMS on NAVSIM v2, establishing a new state-of-the-art.

\section{Related Works}

\subsection{Autonomous Driving World Models}

Vision-centric world models~\cite{babaeizadeh2017stochastic, ha2018recurrent} have gained attention due to their sensor flexibility and data accessibility. Early methods adapted pretrained diffusion models such as Stable Diffusion~\cite{rombach2022high} to driving scenarios but were typically limited to short-term or low-resolution generation and lacked integrated planning capability~\cite{gao2023magicdrive}. 
Recent approaches utilize video generation as the core simulation component. GAIA-1~\cite{hu2023gaia} adopts autoregressive scene generation, while DriveDreamer~\cite{wang2024drivedreamer} and MagicDrive~\cite{gao2023magicdrive} condition diffusion models on BEV maps and 3D bounding boxes. DriveArena~\cite{yang2025drivearena} and DrivingSphere~\cite{yan2025drivingsphere} further advance closed-loop simulation by treating world models as action-conditioned simulators. Voxel-based world modeling approaches explore 3D geometry and spatio-temporal dynamics~\cite{zheng2024occworld}. Large foundation models such as Cosmos~\cite{agarwal2025cosmos} achieve high realism but are computationally expensive.
Despite these advances, existing world models still struggle with high-frequency RL training due to multi-step diffusion inference latency and vulnerability to hallucinations under out-of-distribution actions. Our work addresses these issues by reducing sampling complexity and introducing exploration-constrained reinforcement learning.

\subsection{Reinforcement Learning in World Models}

Reducing real-world trial-and-error cost has motivated RL training inside world model imagination spaces. ReSim~\cite{yang2025resim} and OmniNWM~\cite{li2025omninwm} use trajectory-conditioned video synthesis for evaluative feedback. The Dreamer series~\cite{hafner2019dream,hafner2020mastering,hafner2023mastering} performs multi-step latent rollout optimization.
In autonomous driving, RAD~\cite{gao2025rad} employs 3D Gaussian Splatting world modeling, while AD-R1~\cite{yan2025ad} explores RL within occupancy-based representations. However, these methods face limitations in efficiency, annotation dependency, and underutilization of latent world model features.
Our framework performs RL entirely within latent imagination space, generating dense reward signals directly from internal representations to provide efficient and precise policy optimization without requiring explicit 3D supervision.

\section{Method}

\subsection{Overall}
The overall pipeline of DreamerAD consists of two tightly coupled components: \textbf{World Model with Latent Reward Modeling} and \textbf{Reinforcement Learning with Vocabulary Sampling}.

The first component performs imagination-based trajectory evaluation entirely within latent space. As detailed in Section~\ref{sect:shortcut_wm}, the Shortcut Forcing World Model (SF-WM) predicts future scene representations through single-step latent rollouts. The Autoregressive Dense Reward Model (AD-RM) then evaluates these predicted latent states autoregressively to produce step-wise reward signals across eight driving metrics.

The second component optimizes the policy using trajectories sampled from a predefined high-quality vocabulary. As described in Section~\ref{sect:rl}, Gaussian-weighted vocabulary sampling ensures exploration remains within physically plausible trajectory manifolds, preventing world model hallucinations during RL training\cite{jiang2026wovr}.

\begin{figure}[t]
    \centering
    \includegraphics[width=\linewidth]{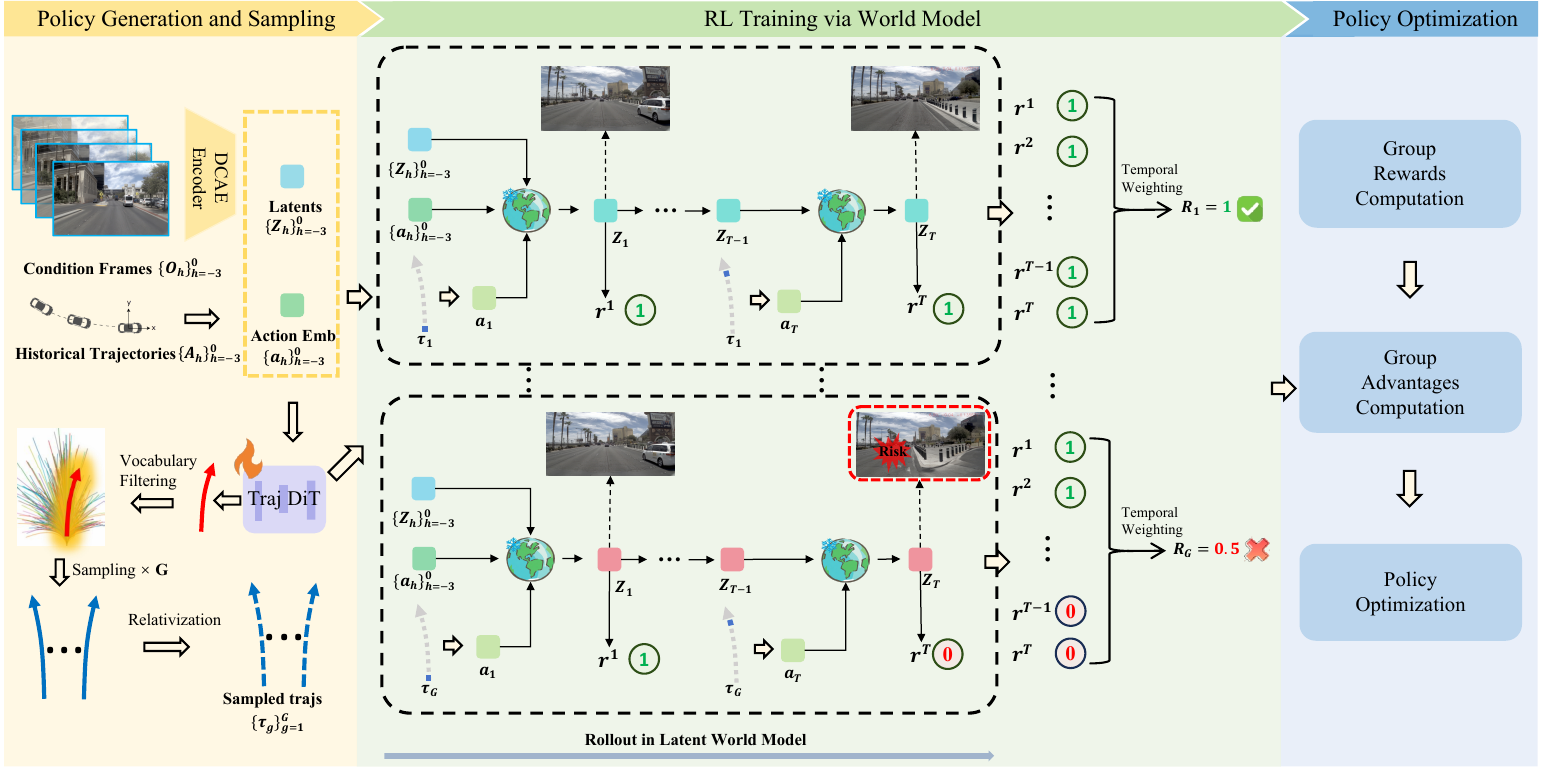}
    \caption{\textbf{Overview of the DreamerAD RL training architecture.} 
    The RL training pipeline consists of three main stages: 1) \textbf{Policy Generation and Sampling (yellow):} Generates a base policy from historical inputs and samples a set of candidate trajectories based on a predefined vocabulary. 2) \textbf{RL Training via World Model (green):} Performs latent rollouts for the sampled trajectories to imagine future states. Step-wise rewards are decoded from these latent features and aggregated into a time-aware dense reward. Notably, our latent representations can be losslessly decoded into RGB frames for accident analysis or visualization, though decoding is bypassed during training for efficiency. 3) \textbf{Policy Optimization (blue):} Computes group advantages from the dense rewards to optimize the policy network using the GRPO algorithm.}
    \label{fig:architecture}
\end{figure}

\subsection{World Model with Latent Reward Modeling}

This section introduces our latent world modeling framework, which is designed to jointly capture future scene dynamics and trajectory evolution. We first present the foundational world model used to construct latent-space predictive representations, then describe the proposed SF-WM that significantly reduces sampling steps while preserving prediction fidelity under low-step inference, and finally describe the AD-RM for trajectory evaluation and latent-space reward modeling.
\subsubsection{Shortcut Forcing World Model}
\label{sect:wm}
\paragraph{\textbf{Foundation World Model.}}
To support imagination-based training, we adopt Epona~\cite{zhang2025epona} as our backbone world model. Epona is an autoregressive diffusion model based on flow matching that unifies video generation and trajectory planning, enabling future video prediction conditioned on action controls.

Given historical observations $O \in \mathbb{R}^{B \times P \times H \times W \times 3}$ and actions $A \in \mathbb{R}^{B \times P \times 3}$, a visual autoencoder and an action encoder compress them into latent embeddings:
\begin{equation}
Z = \text{DCAE-encoder}(O) \in \mathbb{R}^{B \times P \times L \times C}, \quad
a = \text{MLP}(A) \in \mathbb{R}^{B \times P \times 3 \times D}
\end{equation}

Then image embedding $Z$ is processed by a temporal projection module to obtain $Z_{proj} \in \mathbb{R}^{B \times P \times L \times D}$. Concatenating the projected visual tokens and action embeddings along the spatial dimension forms a unified latent representation $E \in \mathbb{R}^{B \times P \times (L+3) \times C}$. 
The final frame of $E$ is used as a compact condition $F \in \mathbb{R}^{B \times (L+3) \times C}$ for the flow matching generator to predict the next-frame latent $\hat{z}_{next} \in \mathbb{R}^{B \times L \times C}$ and the future trajectory $\tau_{pred} \in \mathbb{R}^{B \times T \times 3}$.

The model is jointly trained using the ground-truth next-frame latent $z_{next}$ and future trajectory $\tau_{gt}$. 
Since the original model trained on the NuPlan dataset operates at 10 Hz while the NavSim environment runs at 2 Hz, we first fine-tune the world model on NavSim to adapt to the 2 Hz generation interval before performing step distillation.

\paragraph{\textbf{Shortcut Forcing World Model.}}
\label{sect:shortcut_wm}

Foundation models such as Epona requires 100 sampling steps per frame, creating prohibitive latency for high-frequency RL training. To address this limitation, we propose the Shortcut Forcing World Model (SF-WM), which compresses sampling to 1-4 steps while preserving prediction fidelity—achieving up to $80\times$ faster inference.

Inspired by shortcut models~\cite{frans2024one} and diffusion forcing~\cite{chen2024diffusion}, SF-WM introduces a recursive shortcut forcing mechanism that discretizes the continuous flow process into a multi-resolution step space defined by powers of two. The model is conditioned on both the signal level $t$ and the requested step size $d$ through a step embedding.

Within the rectified flow framework, given conditional latent features $Z$, we define the interpolation
\begin{equation}
x_t = t x_1 + (1-t)x_0,\quad v=\frac{dx_t}{dt}=x_1-x_0.
\end{equation}
where $x_0\sim\mathcal{N}(0,I)$ and $x_1$ denotes the clean data latent representation.

Let $K_{max}$ be the maximum sampling steps and $d_{min}=1/K_{max}$. During training, the step size is sampled as
\begin{equation}
d \sim 1/\mathcal{U}(\{1,2,4,8,\dots,K_{max}\}),\quad 
t \sim \mathcal{U}(\{0,d,2d,\dots,1-d\}).
\end{equation}

Training follows a teacher-student distillation scheme. For $d=d_{min}$, the model is trained using the standard flow matching loss. For $d>d_{min}$, two teacher half-steps are used:
\begin{align}
v_1 &= \phi_\theta(x_t,t,d/2), \\
x_{mid} &= x_t + v_1 d/2, \\
v_2 &= \phi_\theta(x_{mid},t+d/2,d/2).
\end{align}

The target velocity is defined as
\begin{equation}
v_{target}=
\begin{cases}
x_1-x_0,& d=d_{min},\\
\text{sg}((v_1+v_2)/2),& \text{otherwise}.
\end{cases}
\end{equation}

The optimization objective is
\begin{equation}
\mathcal{L}(\theta)=
\mathbb{E}_{x_0,x_1,t,d}
\left[
\omega(t)\|\phi_\theta(x_t,t,d)-v_{target}\|^2
\right],
\end{equation}
where $\omega(t)=0.9t+0.1$ balances global structure and local detail preservation.

\begin{figure}[tbp]
    \centering
    
    \begin{subfigure}{\textwidth}
        \centering
        \includegraphics[width=\linewidth]{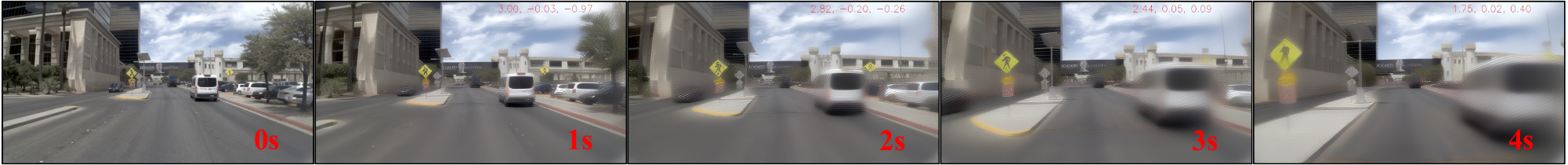}
        \caption{Epona: Cumulative error increases over time, leading to blurry scenes during one-step inference.}
        \label{fig:epona}
    \end{subfigure}

    \begin{subfigure}{\textwidth}
        \centering
        \includegraphics[width=\linewidth]{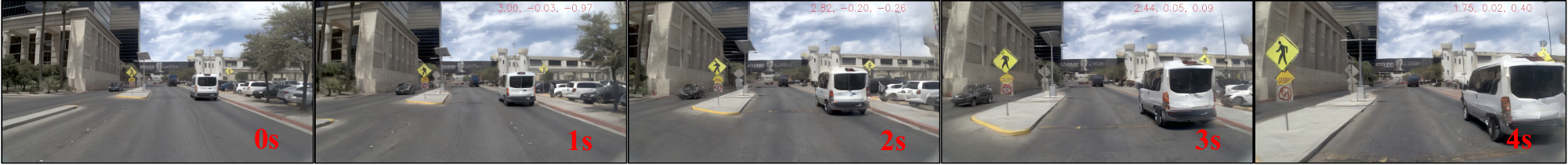}
        \caption{Ours: Visual clarity is maintained across time steps during one-step inference.}
        \label{fig:ours}
    \end{subfigure}
    
    \caption{Visualization of one-step inference for Epona and Shortcut Forcing World Model.} 
    \label{fig:epona vs ours}
\end{figure}

At inference, SF-WM can be conditioned on a desired step size (e.g., $d=1/4$) to generate predictions using only 1–4 sampling steps. As shown in Fig.~\ref{fig:epona vs ours}, SF-WM maintains sharp autoregressive prediction quality under one-step inference, whereas the original model suffers from severe error accumulation and blurring.

\subsubsection{Autoregressive Dense Reward Modeling}
\label{sect:rm}

\paragraph{\textbf{Reward Annotation and Labeling.}}
World models trained on near-expert demonstrations are vulnerable to hallucinations when evaluated on out-of-distribution trajectories with large spatial deviations. To mitigate this issue, we construct a spatially constrained exploration vocabulary.
Specifically, from a large vocabulary of 8192 trajectories, we filter candidates within the neighborhood of human driving trajectories. We extract the end state $(x,y,\theta)$ of each trajectory and compare it with the end state of the corresponding ground-truth trajectory. A trajectory is retained only if it satisfies the lateral and longitudinal constraints $|\Delta y|\le y_{\text{thresh}}$ and $|\Delta x|\le x_{\text{thresh}}$, as well as the heading deviation
\begin{equation}
\Delta \theta = \min(|\theta_{\text{vocab}}-\theta_{\text{gt}}|,\,2\pi-|\theta_{\text{vocab}}-\theta_{\text{gt}}|) \le \theta_{\text{thresh}}.
\end{equation}
We set $x_{\text{thresh}}=10$m, $y_{\text{thresh}}=5$m, and $\theta_{\text{thresh}}=20^\circ$.

To avoid excessive concentration in the candidate set, we further apply a uniform sampling strategy based on lateral offsets. The filtered trajectories are sorted by $|\Delta y|$, and equally spaced samples are selected to obtain $K$ representative trajectories, forming the final vocabulary $\Gamma=\{\tau^0,\tau^1,\dots,\tau^K\}$ with $K=256$. This ensures the reward model observes trajectories with diverse deviation levels.

The filtered trajectories are evaluated in the NavSim PDM simulator to obtain eight reward dimensions 
$r=\{r_{\text{nc}},r_{\text{dac}},r_{\text{ddc}},r_{\text{tlc}},r_{\text{ep}},r_{\text{ttc}},r_{\text{lk}},r_{\text{hc}}\}$.
Unlike prior work that evaluates only full-trajectory scores, we compute rewards under multiple prediction horizons from $0$ to $4.0$s with a step of $0.5$s, producing scores across eight time steps $\{r^1,r^2,\dots,r^8\}$. This enables the reward model to capture both overall trajectory quality and the temporal evolution of rewards, facilitating the trade-off between short-term safety and long-term planning.

\paragraph{\textbf{Reward Model Training.}}
The reward model is parameterized as a neural network that autoregressively predicts trajectory rewards using historical context, formulated as
\begin{equation}
r_{\text{pred}}^t=\text{RewardModel}(\text{traj}_{0:t},\text{his}_{-3:t}),
\end{equation}
where $t\in\{1,\dots,8\}$ denotes the prediction horizon, and $t<0$ represents historical time steps.

During inference, the world model in Section \ref{sect:shortcut_wm} is frozen and autoregressively predicts latent future representations $\{\hat{z}_1,\dots,\hat{z}_t\}$ conditioned on trajectory inputs $\text{traj}_{0:t}$ and latent context $Z$. Historical information is encoded through a multi-layer perceptron:
\begin{equation}
\text{his}_{-3:t}=\text{his\_enc}(\text{concat}[z_{-3},z_{-2},z_{-1},z_0,\hat{z}_1,\dots,\hat{z}_t]).
\end{equation}

Since the latent dimension $L=512$ is high, a learnable query-based compression mechanism reduces it to $l=32$. To distinguish the eight reward dimensions, we initialize eight independent learnable bases $Q_{\text{base}}\in\mathbb{R}^{8\times D}$. Dynamic trajectory and temporal information are encoded as
\begin{equation}
C_{\text{dyn}}=\text{MLP}_{\text{traj}}(\text{traj}_{0:t})+\text{Emb}_{\text{step}}(t),
\end{equation}
and the reward query is defined as
\begin{equation}
Q_r=Q_{\text{base}}+C_{\text{dyn}}.
\end{equation}

Reward representations are decoded through cross-attention followed by an MLP head:
\begin{equation}
r_{\text{pred}}^t=\text{MLP}(\text{Cross-Attention}(Q_r,\text{his}_{-3:t})).
\end{equation}

Training is supervised using binary cross-entropy loss:
\begin{equation}
\mathcal{L}_{\text{sup}}=\sum_{k=1}^{8}\omega_k\cdot\gamma(t)\cdot\text{BCEWithLogits}(r_{\text{pred}},r),
\end{equation}
where $\omega_k$ and $\gamma(t)$ denote reward-type and temporal weighting factors respectively.

\subsection{Reinforcement Learning with Vocabulary Sampling}

\label{sect:rl}
This section presents a reinforcement learning framework for trajectory optimization in latent imagination space. We design a safety-prioritized reward formulation that separates safety compliance and task performance signals, introduce dense temporal reward aggregation to reduce reward sparsity, and adopt Gaussian-guided vocabulary sampling for balanced exploration. Policy optimization is performed using a GRPO-based actor training scheme with behavioral cloning and KL regularization for stable training.

\paragraph{\textbf{Reward Design for RL.}}
Traditional reinforcement learning methods often combine rewards using simple weighted summations, which may neglect the varying importance of different reward components. In autonomous driving, safety is treated as the primary optimization constraint.

Following NavSim~\cite{im2024navsim}, we partition the eight reward dimensions into safety terms $r_{\text{safe}}=\{r_{\text{nc}},r_{\text{dac}},r_{\text{ddc}},r_{\text{tlc}}\}$ and task performance terms $r_{\text{task}}=\{r_{\text{ep}},r_{\text{ttc}},r_{\text{lk}},r_{\text{hc}}\}$. The safety compliance reward is formulated using a log-sigmoid aggregation:
\begin{equation}
L=\sum_{i\in \text{safe}} w_i \log(\text{sigmoid}(r_i)),
\end{equation}
while the task reward is computed as
\begin{equation}
S=\log\left(\sum_{j\in \text{task}} w_j r_j\right).
\end{equation}
The total reward is defined as
\begin{equation}
r_{\text{total}}^t=L+S=\log\left(\prod_i r_i^{w_i}\times \sum_{j} w_j r_j\right),
\end{equation}
for $t\in\{0,1,\dots,7\}$. The logarithmic fusion mechanism ensures that safety violations dominate the reward signal, as collision events drive the safety term toward zero and consequently push $\log(r_i)$ toward negative infinity.

To mitigate reward sparsity in trajectory-level scoring, we introduce step-level dense rewards for temporal credit assignment. Instead of evaluating only full-trajectory outcomes, we retain trajectory quality signals across intermediate prediction horizons. The final reward is computed as
\begin{equation}
r_{\text{final}}=\sum_{t=1}^{8} w_t \cdot r_{\text{total}}^t,
\end{equation}
allowing the model to identify degradation points along the trajectory and better guide optimization.

\paragraph{\textbf{Vocabulary Sampling.}}
Previous stochastic Gaussian exploration methods \cite{yang2025worldrft,zhou2024recogdrive} often suffer from dynamic inconsistency or limited multimodal coverage due to deterministic flow matching sampling. To address this limitation, we propose a Gaussian-based vocabulary sampling strategy to enable more reliable and diverse trajectory exploration. The model first extracts historical and environmental latent representations to generate a baseline trajectory $\tau_{\text{act}} \in \mathbb{R}^{B \times T \times 3}$. Using $\tau_{\text{act}}$ as the mean and a fixed variance $\sigma^2$, we construct a Gaussian distribution:
\begin{equation}
\tau \sim \mathcal{N}(\tau_{\text{act}}, \sigma^2).
\end{equation}
Since the logarithmic Gaussian likelihood is proportional to the negative Mahalanobis distance, trajectory candidates are ranked by computing the Mahalanobis distance between vocabulary trajectories $\Gamma \in \mathbb{R}^{N \times T \times 3}$ and the policy trajectory:
\begin{equation}
d(x,\tau_{\text{act}})=\sum_{t=1}^{T}\sum_{i=1}^{3}\frac{(x_{t,i}-\tau_{\text{act}_{t,i}})^2}{\sigma_{t,i}^2}.
\end{equation}

A mixed sampling strategy is adopted by selecting $g_1$ trajectories according to softmax probabilities for discrimination and $g_2$ trajectories from the Gaussian neighborhood for local exploration, yielding a sampled trajectory set $\tau_{\text{sample}} \in \mathbb{R}^{B \times G \times T \times 3}$ where $G=g_1+g_2$. The sampled trajectories are evaluated by the reward model to obtain final rewards $r^i_{\text{final}}$.

\paragraph{\textbf{Policy Optimization.}}
Policy learning is performed using the GRPO algorithm. The normalized group advantage is computed as
\begin{equation}
A_i=\frac{r^i_{\text{final}}-\text{mean}(r_{\text{final}}^{1..G})}{\sqrt{\text{var}(r_{\text{final}}^{1..G})}}.
\end{equation}
To constrain policy updates, an importance ratio is used:
\begin{equation}
\rho=\exp(\log \pi_\theta-\log \pi_{\text{old}}).
\end{equation}
The actor loss is
\begin{equation}
L_{\text{actor}}=\mathbb{E}\left[\max(-A_i\rho,-A_i\text{clip}(\rho,1-\epsilon,1+\epsilon))\right].
\end{equation}
We further regularize training using behavioral cloning loss $L_{\text{bc}}=\|\tau_{\text{act}}-\tau_{\text{gt}}\|_1$ and KL divergence loss $L_{\text{kl}}=D_{\text{KL}}(\pi_\theta\|\pi_{\text{ref}})$. The final objective is
\begin{equation}
L_{\text{total}}=L_{\text{actor}}+L_{\text{bc}}+L_{\text{kl}}.
\end{equation}

We also evaluate Flow-GRPO~\cite{liu2025flow}, an RL method designed for flow matching architectures, with detailed discussion provided in the supplementary material.

\section{Experiment}

\subsection{Dataset}
We evaluate DreamerAD on the NavSim dataset, which was built upon nuPlan and provides surround-view images from 8 cameras along with high-quality LiDAR point clouds. 
The dataset is split into 1,192 training scenes and 136 testing scenes. During the collection process, static scenarios and constant-speed driving scenarios were excluded to retain highly challenging scenarios. NavSim offers a simulation environment for closed-loop evaluation. NavSim v1 adopts the Predictive Driver Model Score (PDMS) as the evaluation metric, which aggregates multiple driving-related criteria including no collisions (NC), drivable area compliance (DAC), time-to-collision (TTC), comfort (Comf.), and ego progress (EP). 
Building upon this, NavSim v2 introduces the extended PDM Score (EPDMS), incorporating factors such as driving direction compliance (DDC), traffic light compliance (TLC), lane keeping (LK), history comfort (HC), and extended comfort (EC). Metric introduction and calculation method are further detailed in the supplementary material.

\subsection{Implementation Details}
We utilize Epona, which was trained on NuPlan and NuScenes datasets from scratch, as our foundational world model. All images are resized to 512×1024. All training runs are executed on 32 NVIDIA H20 GPUs. 
To facilitate generation and planning on the NavSim dataset, we format the NavSim data and apply the AdamW optimizer with a batch size of 128, a learning rate of $3 \times 10^{-5}$, and a weight decay of $5 \times 10^{-2}$. 
The fine-tuning process spans 5 epochs over approximately one day. 
During the shortcut forcing world model training stage, we train for 12 epochs over three days using identical parameters. 
For reward model training, we apply a batch size of 320 and a learning rate of $3 \times 10^{-4}$ for 12 epochs, completing in about one week. 
In the reinforcement learning phase, we use a batch size of 196 and a learning rate of $1 \times 10^{-4}$ to fine-tune for 2 epochs over approximately 8 hours. 
Inference speed reports are measured on a single NVIDIA H20 GPU. 
We set our VisDiT sampling steps to 1 and TrajDiT sampling steps to 20 across all experiments.

\subsection{Main Results}
As demonstrated in Table~\ref{tab:navsim_v2_comparison}, our method achieves state-of-the-art performance on the NAVSIM v2 closed-loop planning benchmark. Our approach yields an EPDMS of 87.7, outperforming all existing methods and surpassing the Epona baseline by \textbf{2.6} points. This demonstrates that conducting reinforcement learning within a latent space is highly effective. Furthermore, our approach significantly outperforms Epona in critical safety metrics, improving no collisions (NC) by 0.9, time-to-collision (TTC) by 1.1, and drivable area compliance (DAC) by 1.5. These gains demonstrate that imagination-based trial-and-error learning enables the model to acquire robust obstacle avoidance capabilities. The 0.8-point decrease in ego progress (EP) reflects a deliberate safety-first trade-off: prioritizing collision avoidance inherently reduces driving aggressiveness. Additionally, we achieve substantial improvements in lane keeping (LK), history comfort (HC), and extended comfort (EC). These gains indicate that after imagination-based training, the model develops a deeper understanding of driving behaviors and enhances its multi-dimensional driving capabilities.

Furthermore, as shown in Table~\ref{tab:navsim_v1_comparison}, our method achieves a state-of-the-art score of 88.7 among all world-model-based methods on NAVSIM v1. It outperforms the Epona baseline by \textbf{2.5} points overall, with a 2.1 increase in DAC and a 0.5 increase in TTC. These consistent improvements across core safety metrics further validate the effectiveness of our latent imagination training for safe driving. While our overall score is slightly lower than those of AutoVLA and RecogDrive, this is due to differences in the training setup. Those VLA methods rely on more powerful representations from stronger encoders. In contrast, our method achieves highly competitive results using an encoder pre-trained solely on unsupervised driving videos.

In conclusion, our experimental results show that imagination-based reinforcement learning in latent space—driven by extensive trial-and-error interaction—significantly enhances the safety of driving models, demonstrating strong potential for industrial application.
\begin{table*}[t!]
\centering
\caption{\textbf{Comparison with state-of-the-art methods on the NAVSIM v2\cite{im2024navsim} with extended metrics.} 
\textcolor{green!60!black}{+} and \textcolor{red}{-} denote improvement/degradation relative to Epona.}
\label{tab:navsim_v2_comparison}
\resizebox{\textwidth}{!}{
\begin{tabular}{l|c@{\hspace{1em}}c@{\hspace{0.5em}}c@{\hspace{1em}}c@{\hspace{1.2em}}|c@{\hspace{1em}}c@{\hspace{1em}}c@{\hspace{1.3em}}c@{\hspace{1.2em}}c@{\hspace{1.5em}}|>{\columncolor{gray!25}}c}
\toprule
\textbf{Method} & \textbf{NC $\uparrow$} & \textbf{DAC $\uparrow$} & \textbf{DDC $\uparrow$} & \textbf{TLC $\uparrow$} & \textbf{EP $\uparrow$} & \textbf{TTC $\uparrow$} & \textbf{LK $\uparrow$} & \textbf{HC $\uparrow$} & \textbf{EC $\uparrow$} & \textbf{EPDMS $\uparrow$} \\
\midrule
TransFuser\cite{prabhu2021transfuser}     & 96.9 & 89.9 & 97.8 & 99.7 & 87.1 & 95.4 & 92.7 & \textbf{98.3} & 87.2 & 76.7 \\
hydramdp++\cite{li2024hydra}      & 97.2 & 97.5 & 99.4 & 99.6 & 83.1 & 96.5 & 94.4 & 98.2 & 70.9 & 81.4 \\
DriveSuprim\cite{liu2024drivesupreme}     & 97.5 & 96.5 & 99.4 & 99.6 & 88.4 & 96.6 & 95.5 & \textbf{98.3} & 77.0 & 83.1 \\
ipad\cite{zhang2024ipad}            & \textbf{98.7} & 97.8 & 99.1 & \textbf{99.8} & 83.5 & 98.0 & 96.2 & 98.1 & 85.6 & 84.1 \\
ReCogDrive\cite{zhou2024recogdrive}      & 98.3 & 95.2 & \textbf{99.5} & \textbf{99.8} & 87.1 & 97.5 & 96.6 & \textbf{98.3} & 86.5 & 83.6 \\
DiffusionDrive\cite{liao2024diffusiondrive}  & 98.2 & 95.9 & 99.4 & \textbf{99.8} & 87.5 & 97.3 & 96.8 & \textbf{98.3} & \textbf{87.7} & 84.5 \\
DriveVLA-W0\cite{li2025drivevla}     & 98.5 & \textbf{99.1} & 98.0 & 99.7 & 86.4 & \textbf{98.1} & 93.2 & 97.9 & 58.9 & 86.1 \\
\midrule
World4Drive\cite{zheng2025world4drive}     & 97.8 & 96.3 & 99.4 & \textbf{99.8} & 88.3 & 97.1 & \textbf{97.7} & 98.0 & 53.9 & 84.8 \\
WorldRFT\cite{yang2025worldrft}        & 97.8 & 96.5 & \textbf{99.5} & \textbf{99.8} & 88.5 & 97.0 & 97.4 & 98.1 & 69.1 & 86.7 \\
\bottomrule
Epona (Base)\cite{zhang2025epona}    & 97.1 & 95.7 & 99.3 & 99.7 & \textbf{88.6} & 96.3 & 97.0 & 98.0 & 67.8 & 85.1 \\
\textbf{Ours}   & 
98.0\up{0.9} & 
97.2\up{1.5} & 
\textbf{99.5}\up{0.2} & 
\textbf{99.8}\up{0.1} & 
87.8\down{0.8} & 
97.4\up{1.1} & 
97.5\up{0.5} & 
\textbf{98.3}\up{0.3} & 
72.4\up{4.6} & 
\textbf{87.7}\up{2.6} \\

\bottomrule
\end{tabular}
}
\end{table*}

\begin{table*}[t!]
\centering
\caption{\textbf{Comparison with state-of-the-art methods on the NAVSIM v1\cite{im2024navsim}.} 
\textcolor{green!60!black}{+} and \textcolor{red}{-} denote improvement/degradation relative to Epona.}
\label{tab:navsim_v1_comparison}
\resizebox{0.95\textwidth}{!}{
\begin{tabular}
{l|@{\hspace{1em}}c@{\hspace{1em}}c@{\hspace{1em}}|c@{\hspace{0.5em}}c@{\hspace{1em}}|c@{\hspace{0.5em}}c@{\hspace{1em}}c@{\hspace{1.5em}} | >{\columncolor{gray!25}[0em][0.5em]}c@{\hspace{0.5em}}}
\toprule
\textbf{Method} & \textbf{Venue} & \textbf{Input} & \textbf{NC $\uparrow$} & \textbf{DAC $\uparrow$} & \textbf{TTC $\uparrow$} & \textbf{Comf $\uparrow$} & \textbf{EP $\uparrow$} & \textbf{PDMS $\uparrow$} \\
\midrule
VADV2\cite{chen2024vadv2}           &arXiv 2024 & C\&L & 97.2 & 89.1 & 91.6 & \textbf{100.0} & 76.0 & 80.9 \\
UniAD\cite{hu2023planning}           &CVPR 2023 & C\&L & 97.8 & 91.9 & 92.9 & \textbf{100.0} & 78.8 & 83.4 \\
TransFuser\cite{prabhu2021transfuser}     &IEEE TPAMI & C\&L & 97.7 & 92.8 & 92.8 & \textbf{100.0} & 79.2 & 84.0 \\
PARA-Drive\cite{weng2024para}     &CVPR 2024 & C\&L & 97.9 & 92.4 & 93.0 & 99.8 & 79.3 & 84.0 \\
DRAMA\cite{yuan2024drama}          &arXiv 2024 & C\&L & 98.0 & 93.1 & 94.8 & \textbf{100.0} & 80.1 & 85.5 \\
Hydra-MDP\cite{li2024hydra}      &arXiv 2024 & C\&L & 98.3 & 96.0 & 94.6 & \textbf{100.0} & 78.7 & 86.5 \\
WOTE\cite{li2025end}           &ICCV 2025 & C\&L & 98.5 & 96.8 & 94.9 & 99.9 & 81.9 & 88.3 \\
DriveVLA-W0\cite{li2025drivevla}    &NeurIPS 2025 & C\&L & \textbf{98.7} & 96.2 & 95.5 & \textbf{100.0} & 82.2 & 88.4 \\
AutoVLA\cite{zhou2025autovla}        &NeurIPS 2025 & C-Only & 98.4 & 95.6 & \textbf{98.0} & 99.9 & 81.9 & 89.1 \\
RecogDrive\cite{zhou2024recogdrive}     &NeurIPS 2025 & C-Only & 97.9 & \textbf{97.3} & 94.9 & \textbf{100.0} & \textbf{87.3} & \textbf{90.8} \\
\midrule
World4Drive\cite{zheng2025world4drive}    &ICCV 2025 & C-Only & 97.4 & 94.3 & 92.8 & \textbf{100.0} & 79.9 & 85.1 \\
WorldRFT\cite{yang2025worldrft}       &AAAI 2026 & C-Only & 97.8 & 96.8 & 94.0 & \textbf{100.0} & 81.7 & 87.8 \\
\bottomrule
Epona (Base)\cite{zhang2025epona}   &ICCV 2025 & C-Only & 97.9 & 95.1 & 93.8 & 99.9 & 80.4 & 86.2 \\
\textbf{Ours} &- & C-Only & 
98.0\up{0.1} & 
97.2\up{2.1} & 
94.3\up{0.5} & 
\textbf{100.0}\up{0.1} & 
83.1\up{2.7} & 
88.7\up{2.5} \\
\bottomrule
\end{tabular}
}
\end{table*}

\subsection{Ablation Studies}


\begin{table}[t]
    \centering
    \small
    \setlength{\tabcolsep}{0.01\linewidth}
    \caption{Ablation study of each component. SF-WM: Shortcut Forcing World Model, AD-RM: Autoregressive Dense Reward Model, RL-SM: RL Sampling Method. One-step inference is used by default during training unless otherwise specified.}
    \resizebox{0.8\linewidth}{!}{
    \begin{tabular}{c|c|c|c|c|c|c}
        \toprule
        \multirow{2}{*}{ID} & \multirow{2}{*}{SF-WM} & \multirow{2}{*}{AD-RM} & \multicolumn{3}{c|}{RL-SM} & \multirow{2}{*}{EPDMS $\uparrow$} \\
        \cline{4-6}
        & & & Vocab Sampling & WorldRFT & Flow-GRPO & \\ 
        \midrule
        1 & & & & & & 85.1 \\
        2 & & $\surd$ & $\surd$ & & & 86.4 \\
        3 & $\surd$ & & $\surd$ & & & 87.0 \\
        4 & $\surd$ & $\surd$ & $\surd$ & & & \textbf{87.7} \\
        5 & $\surd$ & $\surd$ & & $\surd$ & & 86.6 \\
        6 & $\surd$ & $\surd$ & & & $\surd$ & 87.0 \\
        \bottomrule
    \end{tabular}
    }
    \label{tab:component_ablation}
\end{table}

\subsubsection{Impact of Shortcut Forcing Method}
As shown in Table~\ref{tab:component_ablation}, ID 1 represents the Epona baseline. Comparing ID 2 (without Shortcut Forcing) and ID 4 (our full method) demonstrates that Shortcut Forcing (SF) significantly improves driving performance, indicating that SF successfully enhances generation quality even under extreme step compression. 

Furthermore, as detailed in Table~\ref{tab:wm_inference_steps}, single-step inference utilizing our SF method achieves an EPDMS score of 87.7 with an ultra-low latency of just 0.03s. This performance is highly competitive with the 16-step and 4-step settings while operating at a fraction of the time cost. This proves that compressing sampling steps does not compromise downstream policy planning. Ultimately, our single-step inference provides sufficiently rich and robust representations to fully support imagination-based reinforcement learning.

\begin{table}[t]
    \centering
    \begin{minipage}[t]{0.45\textwidth}
        \centering
        \caption{Ablation study of Shortcut Forcing world model inference steps.}
    \begin{tabular}{c|cc}
        \toprule
        Steps & Latency/Frame (s) $\downarrow$ & EPDMS $\uparrow$ \\ 
        \midrule
        16 & 0.40 & 87.7 \\
        4 & 0.10 & \textbf{87.8} \\
        1 & \textbf{0.03} & 87.7 \\
        \bottomrule
    \end{tabular}
    \label{tab:wm_inference_steps}
    \end{minipage}
    \hfill
    \begin{minipage}[t]{0.47 \textwidth}
        \centering
       \caption{Ablation study of reward model training data scale.}
       \resizebox{0.7\linewidth}{!}{
    \begin{tabular}{l|c}
        \toprule
        Data Scale & EPDMS $\uparrow$ \\ 
        \midrule
        Epona Baseline & 85.1 \\
        20\% Training Data & 87.5 \\
        40\% Training Data & 87.5 \\
        100\% Training Data & \textbf{87.7} \\
        \bottomrule
    \end{tabular}
    }
    \label{tab:data_ablation}
    \end{minipage}
\end{table}

\subsubsection{Effectiveness of Autoregressive Dense Reward Model.}
As shown in Table~\ref{tab:component_ablation}, comparing IDs 3 and 4 proves the effectiveness of the autoregressive Dense Reward Model (AD-RM), as it provides crucial temporal-grained reward signals.

To further evaluate its robustness, we trained the reward model using different proportions of the training dataset under identical configurations. As shown in Table~\ref{tab:data_ablation}, the results show very little difference between using 100\% of the data and just 20\% of the data. This indicates that our reward model and training framework successfully learn the essential differences between good and bad driving behaviors purely based on the future states imagined by the world model. It demonstrates that our AD-RM has strong generalization capabilities and broad application potential, requiring only a small amount of data to yield a robust reward signal.

\subsubsection{Effectiveness of Vocab Sampling Method}
As shown in Table~\ref{tab:component_ablation}, IDs 4, 5, and 6 compare three different reinforcement learning sampling methods. ID 4 uses our proposed vocabulary-based sampling, ID 5 uses WorldRFT~\cite{yang2025worldrft}, and ID 6 uses Flow-GRPO~\cite{liu2025flow}. 

WorldRFT causes severe dynamic discontinuity in the sampled trajectories. For a world model that requires high dynamic accuracy, this amplifies hallucinations and increases prediction bias in the reward model. Flow-GRPO attempts to force exploration by directly altering the deterministic Ordinary Differential Equation (ODE) sampling process of flow matching into a Stochastic Differential Equation (SDE) process. This creates a mismatch with the flow training mode and still results in jagged trajectories, though its strong denoising ability makes it perform slightly better than WorldRFT. In contrast, our vocabulary-based Gaussian sampling method generates high-quality trajectories with complete, smooth dynamics. It perfectly complements our autoregressive dense reward model and achieves the best overall performance.

\subsection{Qualitative Results}
\begin{figure}[t]
    \centering
    \includegraphics[width=0.7\linewidth]{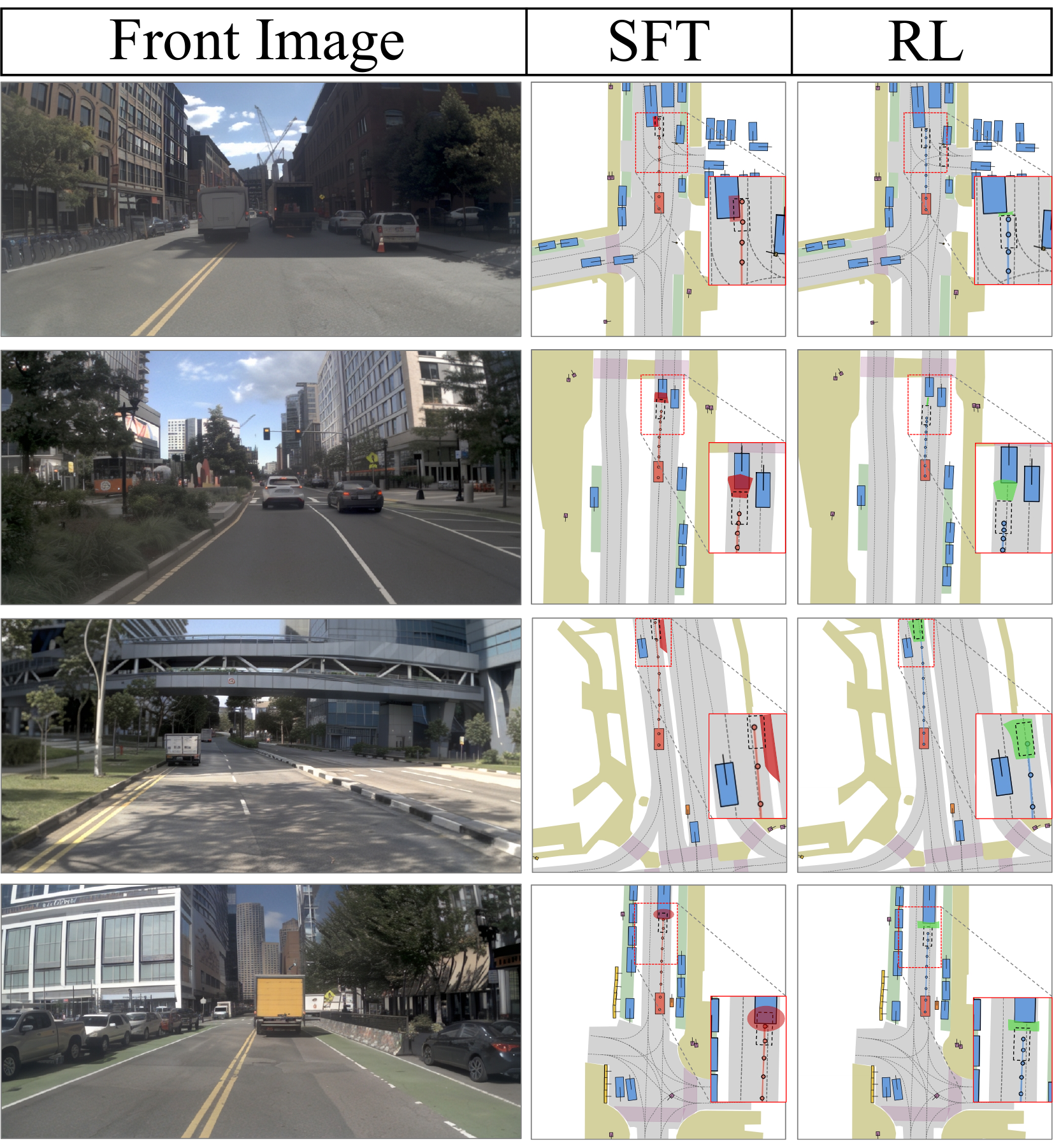}
    \caption{Comparison before and after RL training. The leftmost column displays the front-view camera image at the current timestep. The right columns show the BEV planning results from SFT and RL, respectively. The red trajectory represents the SFT output, while the blue represents the RL output. Red highlights in the SFT BEV maps indicate collisions, whereas green highlights in the RL BEV maps denote safe passage. }
    \label{fig:sft_vs_rl}
\end{figure}

In this section, we qualitatively evaluate DreamerAD on the NavSim dataset. 
As shown in Figure.~\ref{fig:sft_vs_rl}, zooming in on the Bird's-Eye View (BEV) maps in rows 1, 2, and 3 reveals that the Supervised Fine-Tuning (SFT) trajectory maintains an excessively high speed, resulting in collisions with stationary vehicles ahead. Additionally, in row 4, the SFT model collides with the curb. 
Conversely, after Reinforcement Learning (RL) training, the model successfully decelerates and stops appropriately behind the stationary vehicles in rows 1, 2, and 3. In row 4, it correctly adjusts its heading to navigate through safely. This demonstrates that by training within the imagination environment, the model comprehends the severe consequences of poor driving trajectories. Through trial-and-error, it successfully learns safe driving behaviors and accurate decision-making.

\section{Conclusion}
We presented DreamerAD, a framework for reinforcement learning within a visually interpretable latent world model for autonomous driving through three key innovations: shortcut forcing for 80× faster world model inference, autoregressive dense reward modeling for fine-grained credit assignment, and Gaussian vocabulary sampling for physically plausible exploration.
DreamerAD achieves 87.7 EPDMS on NavSim v2, establishing a new state-
of-the-art for closed-loop planning, validating that imagination-based RL training in latent space can effectively learn safe driving behaviors without real-world trial-and-error, opening new avenues for scalable autonomous driving policy optimization.

%
%
\bibliographystyle{splncs04}
\bibliography{main}
\end{document}